\documentclass{article}

\usepackage{arxiv}

\usepackage[utf8]{inputenc} 
\usepackage[T1]{fontenc}    
\usepackage{hyperref}       
\usepackage{url}            
\usepackage{booktabs}       
\usepackage{amsfonts}       
\usepackage{nicefrac}       
\usepackage{microtype}      
\usepackage{lipsum}
\usepackage{graphicx}
\graphicspath{ {./figures/} }

\usepackage{amsthm,amsmath,amssymb}
\usepackage{mathrsfs}
\RequirePackage{subcaption}
\usepackage{setspace}

\title{End2end-ALARA: Approaching the ALARA Law in CT Imaging with End-to-end Learning}

\author{
 Xi Tao \\
  College of Computer Science and Technology\\
  Guizhou University\\
  Guiyang, Guizhou, China 550025 \\
  \texttt{xtao@gzu.edu.cn} \\
   \And
 Liyan Lin \\
  College of Computer Science and Technology\\
  Guizhou University\\
  Guiyang, Guizhou, China 550025 \\
  \texttt{liyanlinn@gmail.com} \\
}

\begin{document}
\maketitle
\begin{abstract}
Computed tomography (CT) examination poses radiation injury to patient.
A consensus performing CT imaging is to make the radiation dose
as low as reasonably achievable, i.e. the ALARA law.
In this paper, we propose an end-to-end learning framework,
named End2end-ALARA,
that jointly optimizes dose modulation and image reconstruction
to meet the goal of ALARA in CT imaging.
End2end-ALARA works by building a dose modulation module
and an image reconstruction module, connecting these modules
with a differentiable simulation function,
and optimizing the them with a constrained hinge loss function.
The objective is to minimize radiation dose subject to
a prescribed image quality (IQ) index.
The results show that End2end-ALARA is able to preset
personalized dose levels to gain a stable IQ level across patients,
which may facilitate image-based diagnosis and downstream model training.
Moreover, compared to fixed-dose and
conventional dose modulation strategies,
End2end-ALARA consumes lower dose to reach the same IQ level.
Our study sheds light on a way
of realizing the ALARA law in CT imaging.
\end{abstract}


\section{Introduction}
Along with its increasing use in clinical diagnosis and treatment,
the ionizing radiation bound with computed tomography (CT) imaging
has been one of the primary concern in the field.
A consensus between practitioners is that the dose used for CT examination
should follow the ALARA law, namely as low as reasonably achievable \cite{GuestLDCT}.
This means the radiation dose should be as low as possible while
the image quality (IQ) meet the requirement of clinical task to be performed.

For a long time, researchers and manufacturers mainly focus on improving
the efficiency of given dose levels
through prefiltering the soft X-rays \cite{TinFiltering},
modulating dose distribution along exposure angles \cite{TCM},
developing high performance reconstruction algorithms \cite{DeepReconSI}, etc.
Such techniques have brought great potential of
radiation dose reduction in the past years.
For example, a clinical study reported that
deep learning reconstruction shows better
lung nodule detection in ultra-low-dose chest CT examination \cite{DLRChest}.
However, those studies often consider uniform dose-level
or dose-reduction-fold across patients.
According to the ALARA law,
the minimum radiation dose for a patient should be predetermined
once the minimum acceptable IQ
for a specific task is expected to be acquired.

\begin{figure}[thbp]
\centering
\includegraphics[width=1.0\textwidth]{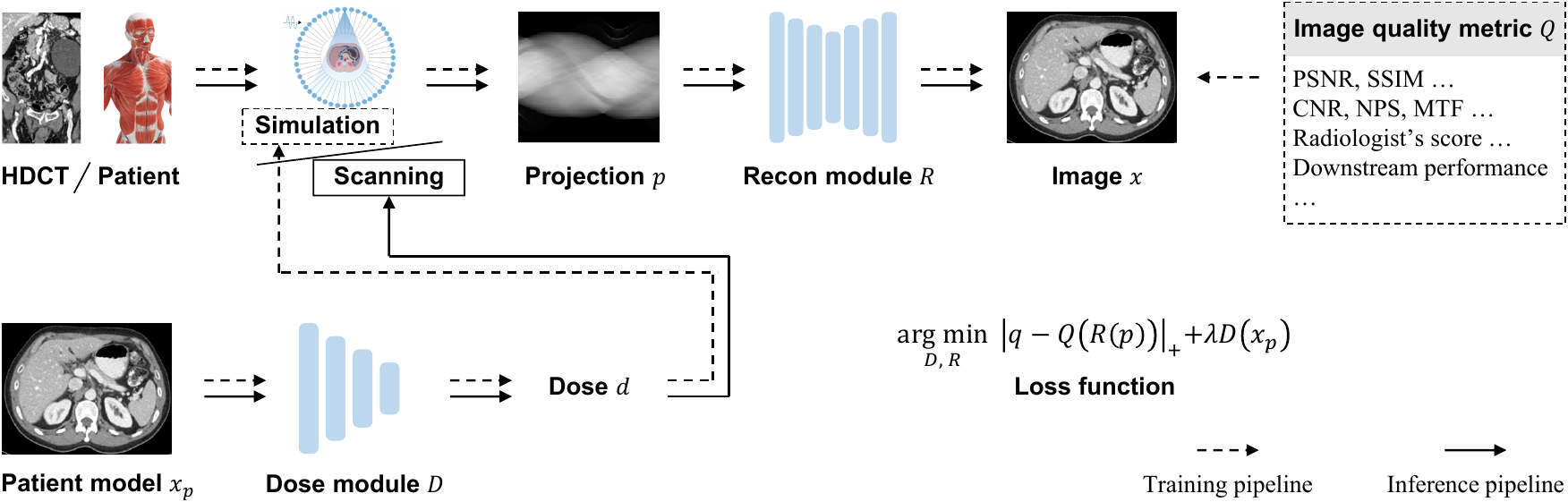}
\caption{The proposed End2end-ALARA framework exploring the way
         to realize the ALARA law in CT imaging.
         The key is to jointly optimize dose
         and reconstruction modules to minimize dose for every patient
         subject to a prescribed quality level $q$.
         }
\label{FigFramework}
\end{figure}

Currently, radiation dose in CT examination is predetermined
by the tube current modulation (TCM) technique \cite{TCM}.
The basic idea is that the number of X-ray photons
reaching the detector bin usually follow Poisson distribution.
Then noise magnitude of projection signal across objects with different anatomies
can be equalized through controlling the incident photon flux with TCM.
This technique could automatically set a proper dose level for each of the patient
subject to a prescribed noise level, in the name of, for example,
\textit{effective mAs} on Siemens's and \textit{noise index} on GE's machine.
However, the technique does not meet the goal of ALARA for two reasons.
Firstly, the control of noise of the projection does not transfer to
that of the image once nonlinear low-dose imaging techniques are employed,
e.g. iterative reconstruction and deep learning reconstruction.
Secondly, the noise magnitude apparently cannot fully reveal the true IQ
perceived by radiologist or required by specific tasks.

In this paper,
we propose an end-to-end learning framework named End2end-ALARA 
to explore the way approaching the ALARA law in CT imaging.
In End2end-ALARA, dose modulation and image reconstruction
are jointly optimized to minimize the radiation dose
for all patient subject to a prescribed IQ level.
This is done by predicting the dose from prior patient model using
a dose module, connecting dose
and reconstruction modules with a differentiable simulation function,
and supervising the them
with a constrained hinge loss function,
as shown in Fig. \ref{FigFramework}.
The experimental results show that,
comparing with the fixed-dose and conventional TCM methods,
our method is able to preset a proper dose level for every single image case
subject to the preset IQ value.
This means End2end-ALARA produces CT images with stable IQ
across patients,
which might reduce the fluctuation of image interpretation.
Moreover, benefiting from the joint optimization mechanism of
dose modulation and image reconstruction,
End2end-ALARA acquires lower dose level with the prescribed IQ
among competitors.

\section{Materials and Methods}

\subsection{The Proposed Framework}
Figure \ref{FigFramework} illustrates the proposed End2end-ALARA framework.
The main body of this framework comprises a dose module
predicting dose level from the prior patient model
and a reconstruction module producing image from the projection.
The success of the framework relies on joint optimization of
the dose and the reconstruction module,
which is accomplished by minimizing dose
subject to a prescribed image quality level,
and formulated as
\begin{equation}
\label{EqObjective}
\begin{array}{l}
\underset{D,R}{\mathop{\arg\min}} \ D(x_p) \\
s.t. \quad Q(R(p)) \geq q,
\end{array}
\end{equation}
where $D$ and $R$ represent
dose and reconstruction module respectively,
$x_p$ is the prior patient model,
$Q$ represents the function of IQ metric,
$p$ is the projection
and $q$ is the prescribed IQ value.
To ease the training of the proposed framework,
we implement the objective in Eq. \ref{EqObjective}
as a constrained hinge loss and formulate it as
\begin{equation}
\label{EqLoss}
\underset{D,R}{\mathop{\arg\min}} \
|q-Q(R(p))|_+ + \lambda D(x_p),
\end{equation}
where $|\cdot|_+$ represents non-negativity constraint
and $\lambda$ is a positive hyper parameter.

When the quality of the reconstructed image is smaller than $q$,
the loss shown in Eq. \ref{EqLoss} impels
the reconstruction module to maximize image quality
and the dose module to minimize radiation dose.
Once the quality index is larger than $q$,
the reconstruction module stops learning
and the dose module keeps minimizing dose.
With such a mechanism, the dose and reconstruction modules
jointly learn to minimize the radiation dose while
maintaining the output image quality stable at the preset level.

\begin{figure}[thbp]
	\centering
	\includegraphics[width=1\linewidth]{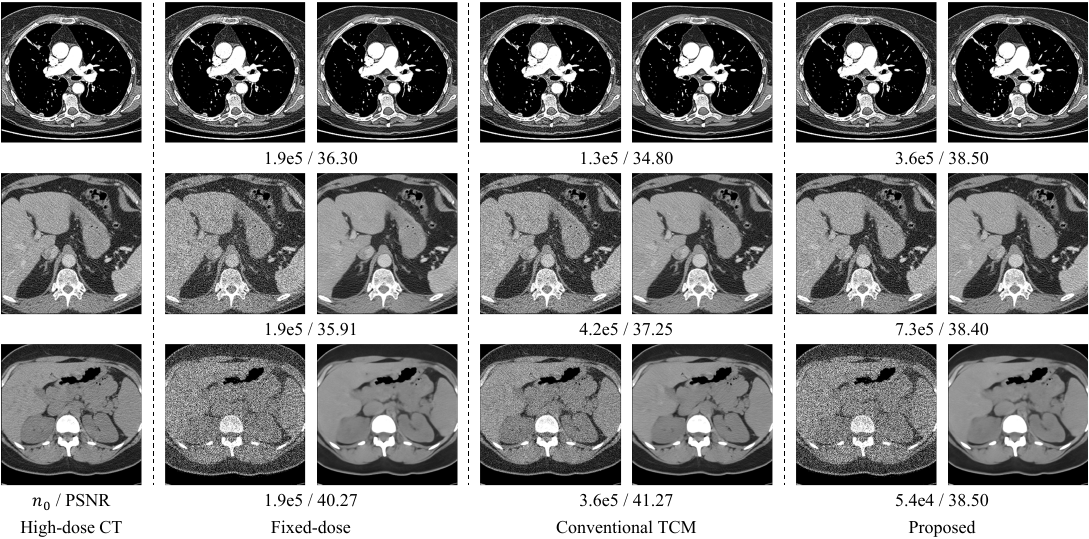}
	\caption{Imaging results by different methods.
             For each method we show FBP-reconstructed and Unet-processed image
             on the left and right respectively.
             The proposed method automatically sets $n_0$ to
             acquire stable PSNRs across patients with various sizes and anatomies.}
	\label{FigImages}
\end{figure}

\subsection{Some Remarks}
To make the training pipeline in the proposed framework executable,
three points should be noted.
Firstly, a pre-scanning patient model is required
to provide prior information for dose prediction.
This can be achieved by scout scanning, i.e. topograms, or prior
unenhanced examination in multi-phase imaging such as perfusion.
Moreover, it is reported that the three dimensional volume
generated from topograms through deep networks is expected
to be a more suitable patient model for dose modulation \cite{ScoutNet}.

Secondly, the simulation procedure needs to be differentiable
to enable gradient to be back-propagated
from the reconstruction module to the dose module.
Considering usual low-dose situation,
the simulation can be performed by simple noise injection,
which is differentiable and formulated as
\begin{equation}
\label{EqNoiseInjection}
p_i = \bar{p}_i + \sigma_i*\eta,
\end{equation}
where $\bar{p}_i$, $p_i$ and $\sigma_i$ represent high-dose
and low-dose projection signal and
standard deviation at $i$th detector bin,
$\eta \sim N(0,1)$ is a random number obeying
standard normal distribution.
There are studies having reported differentiable formulation
of the standard deviation of projection signal.
For example, in Ma's study \cite{VarianceMa}, the variance of projection
in post-log domain is formulated as
\begin{equation}
\label{EqProjVar}
\sigma^2_i = \frac{e^{\bar{p}_i}}{n_{0,i}}
(1+\frac{e^{\bar{p}_i}(\sigma_{e,i}^2-1.25)}{n_{0,i}}),
\end{equation}
where $\sigma_{e}^2$ is the variance of background electronic signal,
and $n_0$ is the incident number of X-ray photons
that represents the radiation dose.

Thirdly, to supervise the proposed framework during training,
a differentiable IQ metric, i.e. function $Q$ in Eq. \ref{EqLoss}, is required.
Actually, a large amount of the full-reference IQ metrics are differentiable
and could be used for optimization of image processing systems.
In order to make the differentiable IQ metrics reveal clinical situation,
their consistency with radiologist's score or downstream task performance
should be verified before employment.
Moreover, radiologist's score or downstream task performance
could be fitted through training deep neural networks \cite{CTIQA},
making the supervision of the proposed framework feasible.

\subsection{Dataset and Implementation}
To verify the effectiveness of End2end-ALARA,
we conducted a preliminary study using the LoDoPaB-CT dataset \cite{LoDoPaBCT},
which contains 42894 images for training, validation and test.
As simple verification, we forward-projected all images to simulate projections with parallel geometry.
For the implementation of our framework,
as the focus of this paper is not acquiring the prior patient model,
we chose the high-dose image as the input of the dose module
with its backbone as a \textit{ResNet-18} network \cite{ResNet}.
The output of the dose module is set as the logarithm
of the number of incident photons, i.e. $n_0=e^d$.
The simulation is performed according to Eqs. \ref{EqNoiseInjection} and \ref{EqProjVar}.
After that, the noisy projection is reconstructed by the filtered backprojection algorithm
and then processed by a \textit{Unet} network \cite{Unet}, forming the reconstruction module.
Lastly, we chose the simple and general PSNR as the IQ metric and prescribed its value as 38.57.

\section{Results}

\begin{figure}[t]
    \captionsetup[subfigure]{labelformat=empty}
	\centering
	\begin{subfigure}{0.33\linewidth}
		\centering
		\includegraphics[width=0.9\linewidth]{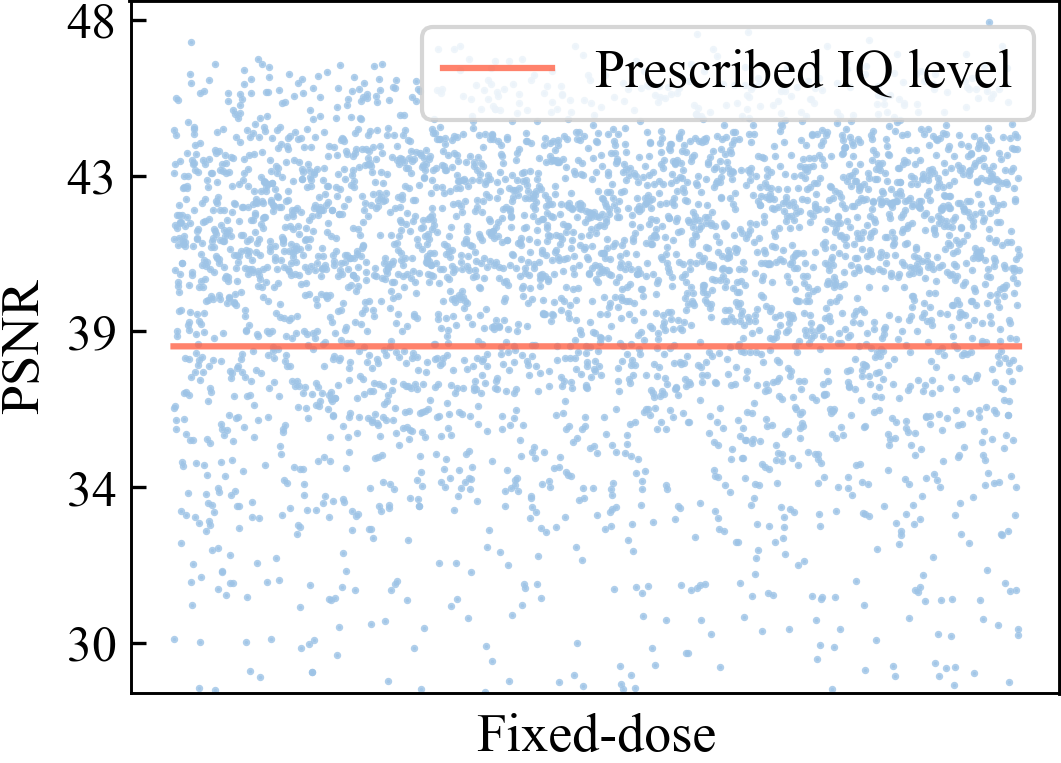}
	\end{subfigure}
	\centering
	\begin{subfigure}{0.33\linewidth}
		\centering
		\includegraphics[width=0.9\linewidth]{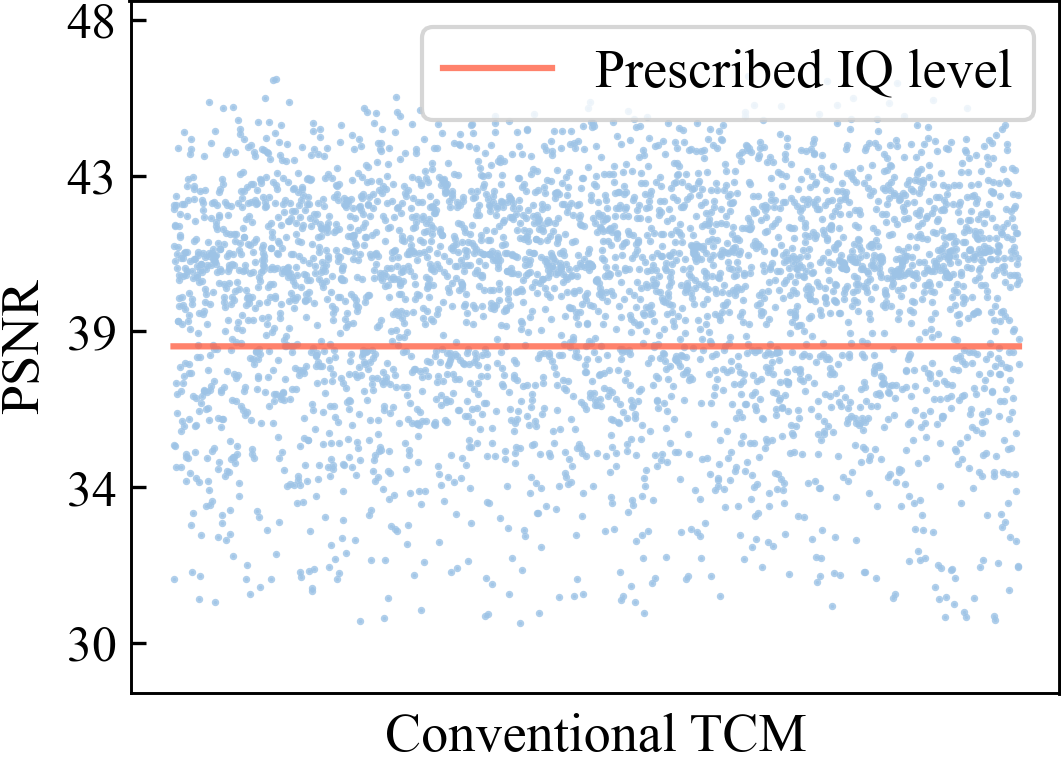}
	\end{subfigure}
	\centering
	\begin{subfigure}{0.33\linewidth}
		\centering
		\includegraphics[width=0.9\linewidth]{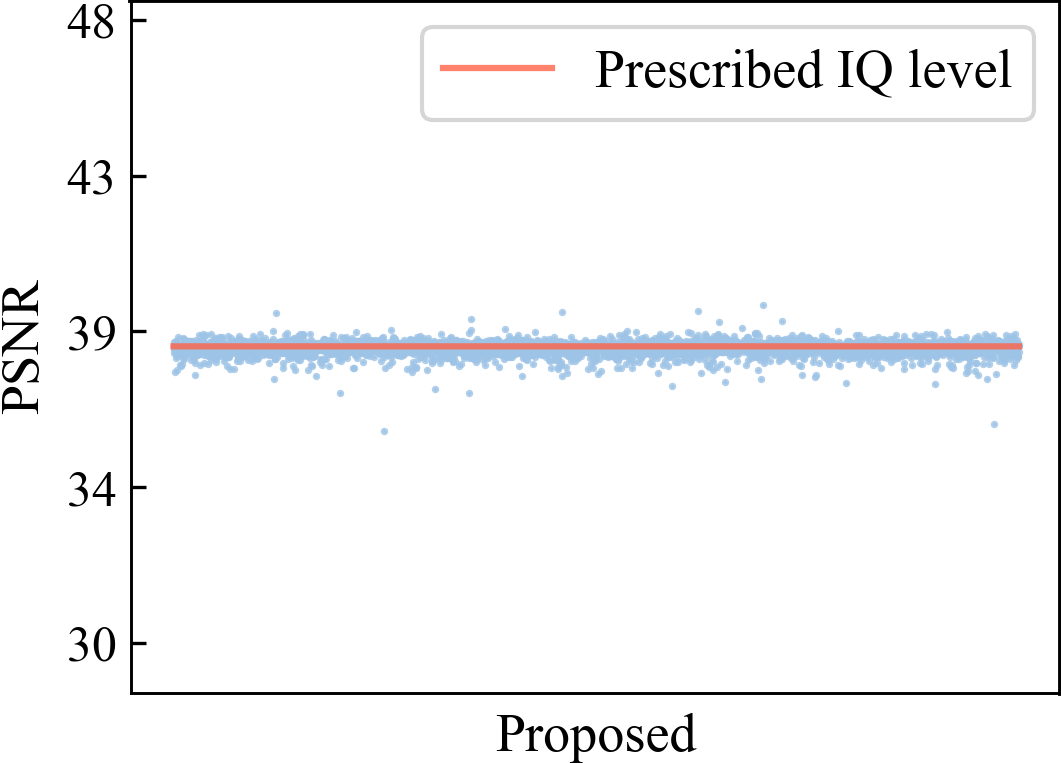}
	\end{subfigure}
	\caption{Scatters of quality indices of 3552 cases in the test set
             by different methods.
             Comparing with the fixed-dose and conventional TCM methods,
             the proposed method produces images with quality stabled at
             the prescribed IQ level.}
	\label{FigQualityScatter}
\end{figure}

Figure \ref{FigImages} shows the imaging results by different methods.
The fixed-dose method uses a constant $n_0$ for all patients.
The conventional TCM method modulates $n_0$ according the attenuation
of each patient to acquire nearly constant detected signal magnitude.
The proposed method learns to modulate $n_0$ to acquire constant image quality.
As can be seen, our method is able to automatically predetermine a proper dose to
acquire consistent IQ across patients with various sizes and anatomies.
This is further demonstrated by the scatters shown in Fig. \ref{FigQualityScatter},
where our method produces images with stable quality for all cases in the test set,
which is unachievable with the other two methods.

\begin{figure}[t]
	\centering
	\hspace{-3mm}\includegraphics[width=0.4\linewidth]{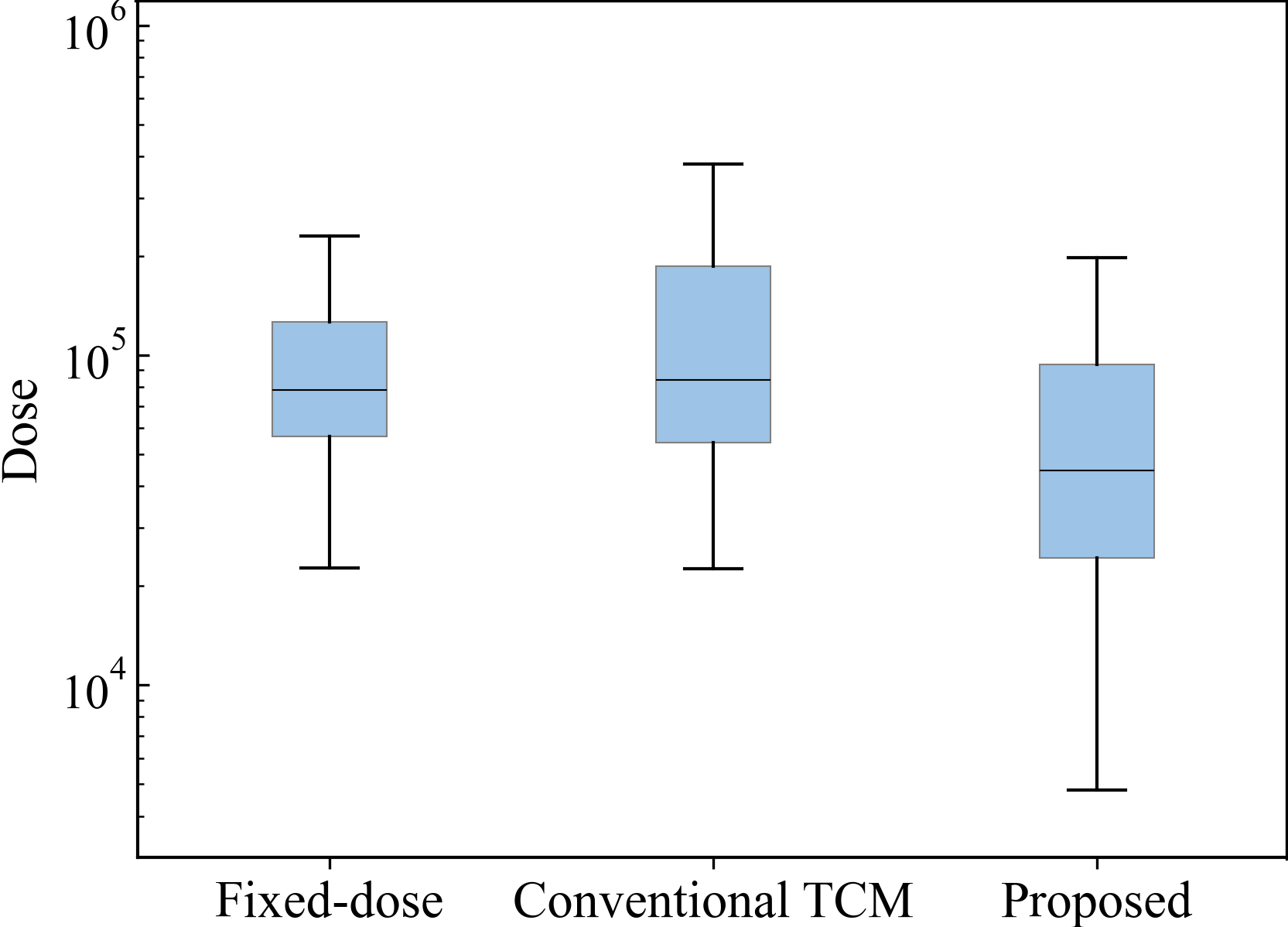}
	\caption{The radiation dose consumed to reach the same IQ level.}
	\label{FigFluxBox}
\end{figure}

Figure \ref{FigFluxBox} shows the radiation dose consumed by different methods
to reach the same IQ level for 340 random cases in the test set.
For all three methods, we preset the PSNR value as 38.57
and then performed a brute force search of dose for
every case to acquire the target IQ.
It is revealed that the proposed method requires the lowest dose among
three competitors, meaning an optimal dose efficiency is gained.
We attribute this merit to the joint optimization mechanism
of dose modulation and reconstruction in the proposed framework.

\section{Conclusion}
In this paper, we proposed an end-to-end learning framework
to jointly optimizing dose modulation and image reconstruction in CT imaging.
Our results demonstrate that the proposed End2end-ALARA framework could achieve
stable image quality with lower dose across patients.
We conclude that our method sheds light to a possible way of
realizing the ALARA law in radiological imaging.
In the future, every part of the proposed framework including
the patient model, the dose module, the simulation function,
the reconstruction module and the IQ metric can be further studied.
We anticipate that the proposed framework could bring new insights to the field.

\bibliographystyle{unsrt}
\bibliography{End2end_ALARA_arXiv}

\begin{thebibliography}{10}

\bibitem{GuestLDCT}
J.~Liang, P.~La~Riviere, G.~El~Fakhri, S.~Glick, and J.~Siewerdsen.
\newblock Guest editorial low-dose ct: What has been done, and what challenges
  remain?
\newblock {\em IEEE Transactions on Medical Imaging}, 36(12):2409--2416, 2017.

\bibitem{TinFiltering}
N.~Saltybaeva, A.~Krauss, and H.~Alkadhi.
\newblock Technical note: Radiation dose reduction from computed tomography
  localizer radiographs using a tin spectral shaping filter.
\newblock {\em Medical Physics}, 46(2):544--549, 2019.

\bibitem{TCM}
M.~Gies, W.~Kalender, H.~Wolf, C.~Suess, and M.~Madsen.
\newblock Dose reduction in ct by anatomically adapted tube current modulation.
  i. simulation studies.
\newblock {\em Medical Physics}, 26(11):2235--2247, 1999.

\bibitem{DeepReconSI}
G.~Wang, M.~Jacob, X.~Mou, Yongyi Shi, and Yonina~C. Eldar.
\newblock Deep tomographic image reconstruction: Yesterday, today, and tomorrow
  - editorial for the 2nd special issue "machine learning for image
  reconstruction".
\newblock {\em IEEE Transactions on Medical Imaging}, 40(11):2956--2964, 2021.

\bibitem{DLRChest}
B.~Jiang, N.~Li, X.~Shi, Shuai Zhang, Jianying Li, Geertruida~H. de~Bock,
  Rozemarijn Vliegenthart, and Xueqian Xie.
\newblock Deep learning reconstruction shows better lung nodule detection for
  ultra-low-dose chest ct.
\newblock {\em Radiology}, 303(1):202--212, 2022.
\newblock PMID: 35040674.

\bibitem{ScoutNet}
J.~Montoya, C.~Zhang, Y.~Li, Ke~Li, and Guang-Hong Chen.
\newblock Reconstruction of three-dimensional tomographic patient models for
  radiation dose modulation in ct from two scout views using deep learning.
\newblock {\em Medical Physics}, 49(2):901--916, 2022.

\bibitem{VarianceMa}
J.~Ma, Z.~Liang, Y.~Fan, Yan Liu, Jing Huang, Wufan Chen, and Hongbing Lu.
\newblock Variance analysis of x-ray ct sinograms in the presence of electronic
  noise background.
\newblock {\em Medical Physics}, 39(7Part1):4051--4065, 2012.

\bibitem{CTIQA}
W.~Lee, F.~Wagner, A.~Galdran, Yongyi Shi, Wenjun Xia, Ge~Wang, Xuanqin Mou,
  Md.~Atik Ahamed, Abdullah Al~Zubaer Imran, Ji~Eun Oh, Kyungsang Kim, Jong~Tak
  Baek, Dongheon Lee, Boohwi Hong, Philip Tempelman, Donghang Lyu, Adrian
  Kuiper, Lars {van Blokland}, Maria~Baldeon Calisto, Scott Hsieh, Minah Han,
  Jongduk Baek, Andreas Maier, Adam Wang, Garry~Evan Gold, and Jang-Hwan Choi.
\newblock Low-dose computed tomography perceptual image quality assessment.
\newblock {\em Medical Image Analysis}, 99:103343, 2025.

\bibitem{LoDoPaBCT}
J.~Leuschner, M.~Schmidt, D.~Baguer, and P.~Maass.
\newblock Lodopab-ct, a benchmark dataset for low-dose computed tomography
  reconstruction.
\newblock {\em Scientific Data}, 8(1):109, 2021.

\bibitem{ResNet}
K.~He, X.~Zhang, S.~Ren, and J.~Sun.
\newblock Deep residual learning for image recognition.
\newblock In {\em CVPR}, pages 770--778, 2016.

\bibitem{Unet}
O.~Ronneberger, P.~Fischer, and T.~Brox.
\newblock U-net: Convolutional networks for biomedical image segmentation.
\newblock In {\em MICCAI}, pages 234--241, 2015.

\end{thebibliography}

\end{document}